\pdfoutput=1
\documentclass[11pt, a4paper, logo, copyright, nonumbering]{hku}
\usepackage{natbib}

\usepackage[utf8]{inputenc} 
\usepackage[T1]{fontenc}    
\usepackage{booktabs}       
\usepackage{amsfonts}       
\usepackage{nicefrac}       
\usepackage{microtype}      
\usepackage{xcolor}         

\usepackage{graphicx}
\usepackage{multirow}
\usepackage{enumitem}
\usepackage{subcaption}
\usepackage{xspace}
\usepackage{makecell}
\usepackage{tcolorbox}

\usepackage{fontawesome5}  
\usepackage{array}
\definecolor{mydarkblue}{rgb}{0,0.08,0.45}
\definecolor{linkcolor}{rgb}{0.956,0.298,0.235}
\definecolor{citecolor}{HTML}{1976D2}
\hypersetup{
    colorlinks=true,
    linkcolor=citecolor,
    filecolor=magenta,
    urlcolor=cyan,
    citecolor=citecolor,
}

\usepackage{rotating}
\usepackage{enumitem}
\usepackage[capitalize,noabbrev]{cleveref}
\crefname{section}{Sec.}{Secs.}
\Crefname{section}{Section}{Sections}
\Crefname{table}{Table}{Tables}
\crefname{table}{Tab.}{Tabs.}

\definecolor{deemph}{gray}{0.6}

\reportnumber{001}

\renewcommand{\today}{}

\title{\centering SynapseRoute: An Auto-Route Switching Framework on Dual-State Large Language Model}

\author{%
  \textcolor{red}{TBD} \\
}
\author[*]{
\small
Wencheng Zhang$^{1*}$ \quad Shiqin Qiao$^{1*}$ \quad Lingjie Luo$^{1*}$ \quad Yinfeng Li$^2$ \quad Chuanyang Zheng$^3$ \quad Qian Xu$^1$ \quad Meng Li$^1$ \quad Yong Gui$^1$ \quad Yijun He$^1$ \quad Jianing Qiu$^3$ \quad Jindong Hong$^{4\dag}$ \quad Jiankai Sun$^{1\dag}$ \\

\small
$^1$Bytedance \quad $^2$Xidian University \quad $^3$The Chinese University of Hong Kong \quad $^4$Peking University \\
\small
$^*$Equal Contribution \quad $^\dag$Corresponding Authors \\
}

\begin{abstract}
With the wide adoption of large language models (LLMs) in practical applications, selecting an appropriate model requires balancing not only performance but also operational cost. The emergence of reasoning-capable models has further widened the cost gap between “thinking” (high reasoning) and “non-thinking” (fast, low-cost) modes.
In medical benchmarks, question complexity varies substantially, from routine medication adjustments to complex rare disease diagnoses and multidisciplinary consultations. These tasks impose diverse and sometimes conflicting demands on LLMs: while the thinking mode generally excels at complex logical inference, it is associated with overthinking, high computational cost, and a higher rate of hallucination; conversely, the non-thinking mode provides faster, more cost-effective responses for simple queries but struggles with complex reasoning.
In this work, we reveal that approximately 58\% of medical questions can be accurately answered by the non-thinking mode alone, without requiring the high-cost reasoning process. This highlights a clear dichotomy in problem complexity and suggests that dynamically routing queries to the appropriate mode based on complexity could optimize accuracy, cost-efficiency, and overall user experience. Based on this, we further propose SynapseRoute, a machine learning-based dynamic routing framework that intelligently assigns input queries to either thinking or non-thinking modes. Experimental results on several medical datasets demonstrate that SynapseRoute not only improves overall accuracy (0.8390 vs. 0.8272) compared to the thinking mode alone but also reduces inference time by 36.8\% and token consumption by 39.66\%. Importantly, qualitative analysis indicates that over-reasoning on simple queries can lead to unnecessary delays and even decreased accuracy, a pitfall avoided by our adaptive routing. This work further introduces the Accuracy-Inference-Token (AIT) index to comprehensively evaluate the trade-offs among accuracy, latency, and token cost. Across various task scenarios, models equipped with SynapseRoute consistently achieve superior AIT scores, validating the framework’s effectiveness and adaptability in medical contexts.
\end{abstract}
\begin{document}

\maketitle

\section*{Introduction}

In recent years, large language model (LLM) technology has advanced rapidly~\citep{bommasani2021opportunities,firoozi2025foundation,xu2025seqpo}, leading major companies to release two primary types of models: chat-optimized models and reasoning models. The former excel at efficient dialogue, rapid responses, and low-cost interactions. While they possess some inference capabilities, they are limited in handling complex logic and multi-step computational tasks~\citep{chang2024survey}. The latter, conversely, are adept at logical reasoning, multi-step thinking, code comprehension, and abstract task processing, demonstrating superior capabilities when addressing complex problems~\citep{zheng2023lyra,sun2023survey,ye24reasoning}. With the introduction of open-source reasoning models like DeepSeek-R1~\citep{guo2025deepseek}, the ability of open-source models to handle complex problems has significantly improved, approaching or even matching the performance of closed-source proprietary models~\citep{jaech2024openai}. However, while reasoning models offer higher accuracy, they come with a substantial increase in computational and time costs. For instance, GPT-4.1, a chat-optimized model developed by OpenAI, costs \$8 per million tokens, whereas the reasoning model OpenAI o3 is \$40 per million tokens, which is five times more expensive\footnote{\url{https://openai.com/api/pricing/}}.
\begin{figure}
    \centering
    \includegraphics[width=\linewidth,trim={0 5.5cm 0 0},clip]{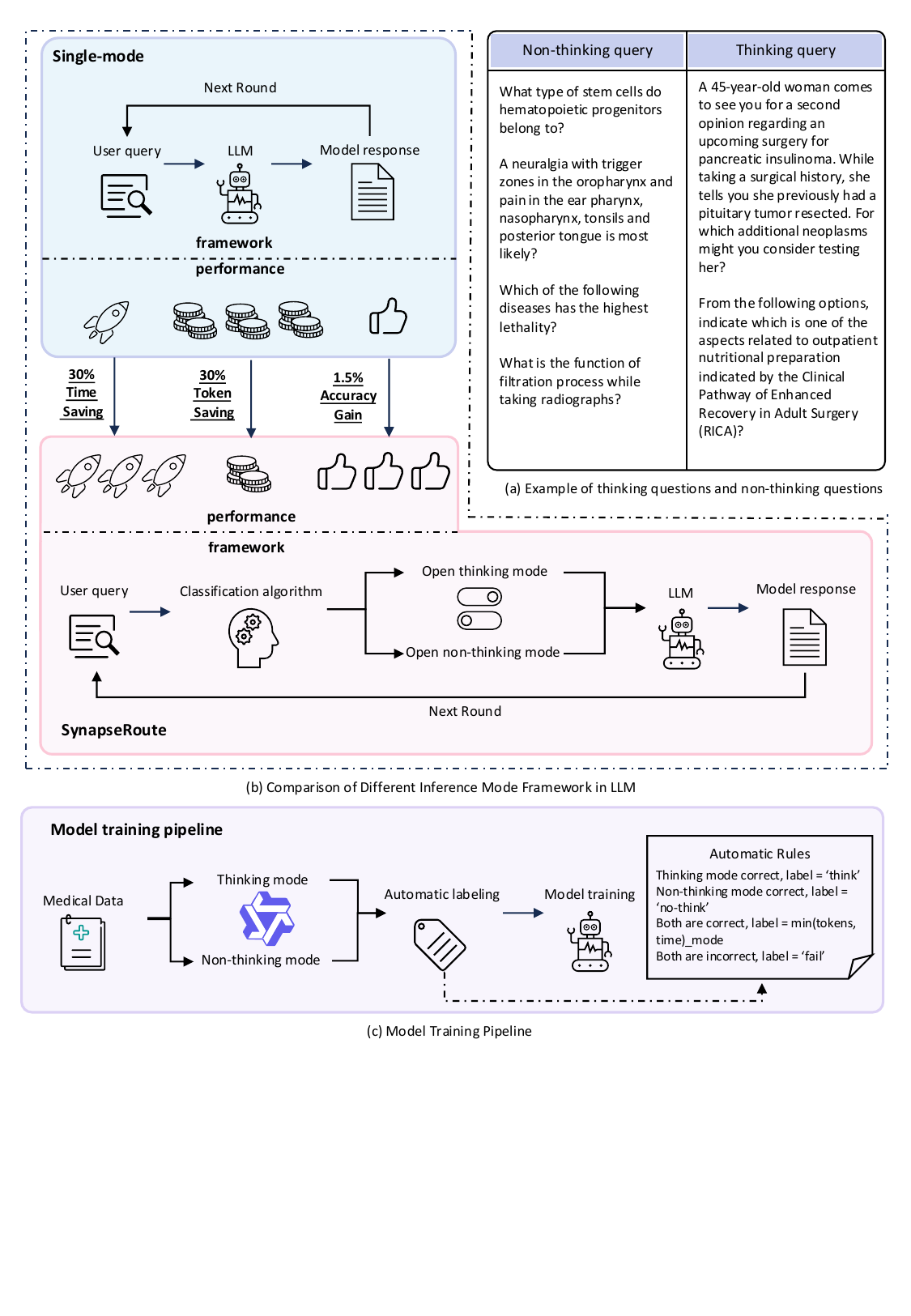}
    \caption{Illustrative examples of thinking and non-thinking questions, comparison of inference mode frameworks in large language models (LLMs), and the proposed model training pipeline.}
    \label{fig:pipeline}
\end{figure}

Currently, some companies are not only releasing these two types of models separately but are also attempting to integrate both modes into a single model. For example, open-source model Qwen3~\citep{yang2025qwen3} supports manual switching of thinking modes within a single model; developers can enable a "thinking mode" for complex problems and use a fast-response "non-thinking mode" for simple tasks. Gemini 2.5 Flash similarly introduces a "thinking mode" switch, allowing developers to flexibly balance quality, cost, and response latency~\citep{team2023gemini}. This trend reveals an important direction: merging both modes into a unified LLM may become one of the mainstream solutions in the future. Compared to deploying two independent LLMs, deploying a single dual-mode LLM offers advantages in system architecture, resource utilization, and operational costs. However, it is crucial to note that the dual-mode architecture still faces cost trade-offs, with the thinking mode being significantly more expensive than the non-thinking mode. For example, with Gemini 2.5 Flash, the thinking mode costs \$3.5 per million tokens, which is 5.8 times higher than that of the non-thinking mode\footnote{\url{https://ai.google.dev/gemini-api/docs/pricing}}. Based on this, a key challenge is how to automatically assess task complexity and intelligently switch between fast and accurate modes within a unified model, maximizing accuracy when needed while minimizing computational cost.

This problem is particularly critical in the medical domain. Medical tasks generally require integrating specialized knowledge with individual patient information, and accuracy demands are extremely high~\citep{qiu2023large,qiu2024application,qiu2024llm,thirunavukarasu2023large,qiu2025emerging}. As LLMs become increasingly powerful and sophisticated, some tasks previously considered to require "thinking" can now be well-answered without enabling a thinking mode. In fact, in daily practice, answers to most medical questions are still based on knowledge (whether it is retrieval or summarization) and do not necessarily require complex reasoning. Over-using the thinking mode could lead to awkward interactions between AI and users, induced by model's tendency to overthink. Therefore, dynamically identifying whether a medical task needs to enable a thinking mode, and subsequently reducing computational resource consumption and inference costs while ensuring accuracy and natural AI-user interactions, becomes valuable and pragmatic approach. This study proposes SynapseRoute, a dynamic routing framework for dual-mode LLMs. As shown in Figure~\ref{fig:pipeline}, compared to a traditional single-mode framework, SynapseRoute automatically determines whether a medical query requires activation of the thinking mode and routes it to the most suitable mode to generate a response. This method analyzes task features and comprehensively considers accuracy, token cost, and time overhead to dynamically make the optimal mode selection decision. Our results show that SynapseRoute significantly reduces response time and token costs on the medical dataset by over 30\% compared to models exclusively using thinking mode. Simultaneously, it achieves an accuracy of 0.8390, surpassing the performance of the thinking mode (0.8272). This cost reduction is attributed to the non-thinking mode's sufficient capability for handling simple tasks, allowing more questions to be processed via a low-overhead path, thereby significantly improving resource utilization efficiency. The enhanced accuracy stems from the dynamic routing framework's ability to intelligently identify and avoid questions where thinking mode might otherwise lead to errors due to reasoning redundancy and hallucinations induced by its over-confidence\footnote{\url{https://treelli.github.io/blog/2025/reasoning-hallucination/}}, thus elevating overall decision quality. We observed that thinking mode errors caused by reasoning redundancy largely occur when questions are direct and information is relatively brief. In these instances, the model tends to automatically associate extraneous background information or even self-defined terminology, leading to inferential inaccuracies and hallucinations. Notably, this routing framework not only holds significant value in the medical domain but also has the potential to be applied to broader domains beyond medicine. By fully leveraging the low-cost efficiency of the non-thinking mode and the deep reasoning capabilities of the thinking mode, SynapseRoute can achieve optimal cost while ensuring accuracy, providing new insights for next-generation LLM architectures and applications.

In summary, this work makes three key contributions in the mechanism design, data annotation methodology, and evaluation of the emerging dual-mode LLMs:

\begin{itemize}
    \item We introduce SynapseRoute, the first intelligent routing system specifically designed for dual-mode LLMs supporting "thinking" and "non-thinking" modes within a unified architecture. This framework automatically identifies the inference complexity of input medical queries based on their characteristics and dynamically switches to the most appropriate inference mode. Our study shows that in typical medical tasks, including complex polysymptomatic diagnoses, multi-disciplinary collaborative assessments, and routine medication adjustments, LLMs operating under the SynapseRoute framework outperform single-mode LLMs. This achieves a significant reduction in token cost and response latency while maintaining high accuracy. Compared to traditional solutions, SynapseRoute substantially simplifies system architecture and enhances the practical utility of LLMs in high-frequency clinical scenarios.
    \item This study proposes an automated data annotation approach that significantly reduces data labeling costs in the medical domain. By integrating accuracy with inference time and token cost metrics, we establish a question type discrimination framework that provides a quantitative basis for distinguishing between "thinking questions" and "non-thinking questions". This approach analyzes the response characteristics of models in both modes to achieve automated annotation of medical question types within datasets, providing the foundational training data for subsequent intelligent routing model inference modes. This innovative attempt overcomes the subjectivity limitations of traditional manual annotation, setting up an annotation mechanism based on algorithmic objectives, and improving the objectivity and efficiency of question type classification.
    \item To comprehensively assess model performance and cost, this research introduces Accuracy-Inference-Token (AIT) index, a novel multi-dimensional evaluation framework for the medical domain, integrating accuracy with time and token costs. This framework quantitatively measures model efficiency and economic viability by evaluating answer accuracy, inference time, and token cost during the response generation process. This aligns more closely with the real-world assessment needs of medical scenarios. Specifically, by establishing a mathematical evaluation model that incorporates accuracy and cost, it systematically analyzes the performance-cost trade-offs across different scenario preferences, providing quantifiable decision support for model optimization. This innovative framework transcends the limitations of traditional single-performance evaluations by incorporating the cost dimension into the model optimization objective, offering a new benchmark for comprehensive performance assessment of LLM efficiency and cost in medical applications.
\end{itemize}

\section*{Results}

We conducted a systematic evaluation of medical question-answer pairs on our test set. We assessed the Qwen3-30B-a3b base LLM across three inference modes: Thinking Mode, Non-thinking Mode, and the Dynamic Mode (SynapseRoute), which is managed by our dynamic routing algorithm. For each question-answer pair, we recorded the model's response content, inference time, and token consumption. This allowed us to compare the distinct performance of these three modes using both Single-Dimensional Performance Metrics and Cross-Dimensional Joint Evaluation Metrics.

\subsection*{Single-Dimensional Performance Metrics Analysis}

In multiple-choice question tasks, the choice of inference mode significantly impacts the final outcome. Generally, thinking mode, relying on multi-turn reasoning and deep semantic analysis, is better suited for problems with complex structures, long logical chains, and strong contextual dependencies, providing more precise judgments. However, the thinking mode's accuracy can exhibit a non-monotonic trend, initially rising and then declining with output length~\citep{su2025between}. This means it might introduce "logical noise" due to "over-deduction," leading to unnecessary errors. This observation aligns with findings in the Qwen3 technical report~\citep{yang2025qwen3}, where researchers also noted that in certain benchmarks, the thinking mode's performance might "slightly degrade" compared to the non-thinking mode~\citep{yang2025qwen3}. In contrast, non-thinking mode, with its lightweight computational path, is more efficient when handling clear and logically simple information but can suffer from broken inference chains in complex tasks, reducing accuracy. The SynapseRoute framework is precisely a dynamic routing solution proposed in this context: it aims to use the model's computational resources "where they matter most" by identifying problem complexity and intelligently switching to the optimal inference path.

Experimental results confirm the effectiveness of this design philosophy. As shown in the Table~\ref{tab:single-dimensional}, thinking mode's accuracy (0.8272) significantly surpasses non-thinking mode (0.5774), further validating the value of deep reasoning mechanisms in complex medical question-answering tasks. However, the SynapseRoute mode further improved accuracy to 0.8390, surpassing the results of exclusively using thinking mode for the first time. This validates the feasibility and effectiveness of the dynamic routing strategy in optimizing overall performance. In terms of specific metrics, SynapseRoute outperforms both thinking and non-thinking modes across Macro Precision (0.8443), Macro Recall (0.8404), and Macro F1 (0.8411), indicating its stronger generalization ability and more balanced performance across various sample categories. Furthermore, it also leads in Weighted Precision (0.8419), Weighted Recall (0.8390), and Weighted F1 (0.8393) when accounting for real sample distribution, demonstrating its stable and superior performance even with imbalanced medical data.

\begin{table}[htbp]
    \centering
    \caption{Comparison of Single-Dimensional Performance Metrics across Inference Modes}
    \label{tab:single-dimensional}
    \begin{tabular}{p{4.2cm} >{\centering\arraybackslash}p{2.8cm} >{\centering\arraybackslash}p{2.8cm} >{\centering\arraybackslash}p{3.6cm}}
        \toprule
        \textbf{Metric} & \textbf{Non-Thinking Mode} & \textbf{Thinking Mode} & \textbf{Dynamic Mode (SynapseRoute)} \\
        \midrule
        Accuracy              & 0.5774 & 0.8272 & 0.8390 \\
        Macro Precision       & 0.4817 & 0.8289 & 0.8443 \\
        Macro Recall          & 0.4561 & 0.8312 & 0.8404 \\
        Macro F1 Score        & 0.4555 & 0.8273 & 0.8411 \\
        Weighted Precision    & 0.6001 & 0.8301 & 0.8419 \\
        Weighted Recall       & 0.5774 & 0.8272 & 0.8390 \\
        Weighted F1 Score     & 0.5720 & 0.8258 & 0.8393 \\
        \bottomrule
    \end{tabular}
\end{table}
While prioritizing accuracy, we also compared SynapseRoute's resource consumption to that of the thinking mode. Our results in Table~\ref{tab:inference_time} show that SynapseRoute's average inference time was 10.8 seconds, significantly lower than the thinking mode's 17.1 seconds, representing a 36.8\% reduction. In terms of token usage, the averages were 476.4 for SynapseRoute and 789.5 for thinking mode, meaning SynapseRoute reduced token consumption by 39.66\%. These findings indicate that SynapseRoute not only boosts accuracy but also effectively cuts down on inference time and token overhead, thereby lowering model usage costs and delivering a more efficient, controlled user experience for real-world applications.

\begin{table}[]
    \centering
    \caption{Comparison of inference time and token size}
    \label{tab:inference_time}
    \begin{tabular}{ccc}
\toprule
& Thinking mode & Dynamic Mode (SynapseRoute) \\
\midrule
Inference time (s) & 17.1071 & 10.8311 \\
Token size & 789.4933 & 476.3598 \\
\bottomrule
    \end{tabular}
\end{table}
These results demonstrate SynapseRoute's core advantage: by intelligently identifying problem complexity and dynamically assigning the most appropriate inference path, the method effectively circumvents the limitations of single-mode models in adapting to diverse tasks. While retaining the high accuracy benefits of thinking mode, SynapseRoute further boosts overall prediction performance, surpassing thinking mode in the accuracy dimension and showcasing superior task matching capability and inference adaptability. 

We further analyzed instances where thinking mode failed due to "overthinking," particularly on questions that were more direct knowledge recall. For example, in Table~\ref{tab:case}, queries like "Mandibular Molar is" or "According to the theory based on mentalization, what characteristic is related to the simulated pre-mentalizing mode?" involve minimal contextual information and are often straightforward inquiries, assuming the responder possesses foundational knowledge.

\begin{table}[]
    \centering
    \caption{Representative cases where failures in the Thinking Mode contrast with successes in the Non-Thinking Mode, due to redundant or inefficient reasoning.}
    \label{tab:case}
    \begin{tabular}{m{2.5cm}m{12.5cm}}
    \toprule
       Question & Redundant Thinking Process \\
       \midrule
Mandibular Molar is & "… But wait, the mandibular molars in humans are typically tricuspid, right? But the question isn't specific to humans, just "mandibular molar" in general …" \\
\hline
According to the theory based on mentalization, what characteristic is related to the simulated pre-mentalizing mode? & "… First, I need to recall what mentalization is. Mentalization refers to the ability to understand and interpret one's own and others' mental states, like thoughts, feelings, and beliefs. It's a key concept in psychotherapy, especially in treatments like Mentalization-Based Treatment (MBT) for personality disorders.

Now, the question is about the ""simulated pre-mentalizing mode."" I'm not entirely sure about the exact terminology here, but I can infer. Pre-mentalizing might refer to a stage before someone can effectively mentalize. So, the simulated aspect could mean that the individual is trying to mentalize but in a way that's not fully developed or is deficient …" \\
\hline
A patient with a history of diabetes for one year with no other complications should have an ophthalmic examination? & "… I remember that diabetes can lead to diabetic retinopathy, which is a common complication. The American Diabetes Association guidelines... I think they recommend that people with type 1 diabetes should have their first dilated eye exam after five years of diagnosis. But wait, the patient here has had diabetes for one year. So maybe the answer isn't B or C. But wait, maybe there's a difference between type 1 and type 2? The question doesn't specify, but maybe it's assuming type 1? Or maybe the general guideline.

Wait, for type 2 diabetes, the recommendation is usually to have an initial eye exam at the time of diagnosis. But the question says the patient has had diabetes for one year. If it's type 2, they might have already had the exam. But if it's type 1, then after five years. …" \\
\bottomrule
    \end{tabular}
\end{table}
Upon detailed dissection of the inference process in thinking mode, we observed that for simple questions like "Mandibular Molar is," the model would sometimes stray beyond the question's scope. It might ponder whether the question pertains to humans, or even begin comparing it to other mammals. During this process, the model would vacillate between the correct answer and a ultimately incorrect one, often settling on the wrong choice with apparent uncertainty. When faced with questions lacking detailed background and context, thinking mode frequently conjured distantly-related contextual information, which interfered with subsequent reasoning. Furthermore, for specialized terminology, the model sometimes defined terms based on its own understanding (e.g., "the simulated pre-mentalizing mode") rather than directly recalling their established meanings.

Overall, errors in thinking mode attributed to overthinking largely stemmed from the questions' succinct nature. The model would automatically associate superfluous background information as a basis for judgment or self-define terminologies, leading to incorrect responses.

\subsection*{Cross-Dimensional Joint Evaluation Metrics Anlysis}

The AIT Index, by definition, ranges from 0 to 1, with higher values indicating a superior balance between model performance and cost. Figure~\ref{fig:ait_by_scenario_0605} below presents the AIT performance of the three inference modes across different evaluation scenarios, with 1,000 bootstrapping iterations enhancing the statistical stability and reliability of the results. In the Figure~\ref{fig:ait_by_scenario_0605}, the horizontal axis progressively reduces the weight of the accuracy metric from right to left, while simultaneously increasing the weight of token cost. This setup simulates real-world application demands with varying emphases on accuracy versus resource trade-offs. Across all comparison scenarios, SynapseRoute's AIT Index consistently outperforms single inference modes, demonstrating its ability and practical value in dynamically balancing precision and efficiency.

Taking the "Accuracy First" scenario as an example, where accuracy is given the highest weight, thinking mode yields an AIT Index of 0.834, while SynapseRoute achieves 0.848, showing a slight lead. In contrast, non-thinking mode scores only 0.619. The root cause of this disparity lies in non-thinking mode's accuracy of just 0.5774, significantly lower than thinking mode (0.8272) and SynapseRoute mode (0.8390). In highly complex tasks like medical question answering, non-thinking mode struggles to provide sufficient reasoning capabilities for high-quality answers, thus severely limiting its performance in the "Accuracy First" scenario. As the weight of accuracy gradually decreases and the weight of token cost correspondingly increases, non-thinking mode shows a slight recovery in its AIT Index due to its low computational overhead. However, since our minimum accuracy weight is still maintained above 0.5, considering the rigid requirement for answer reliability in medical contexts, its overall AIT performance consistently lags behind the other two modes.

Across all scenarios, the gap between SynapseRoute and thinking mode, though not vast, remains consistently in SynapseRoute's favor. This is primarily attributable to SynapseRoute's ability to intelligently avoid errors caused by "over-reasoning," leading to a more precise mode selection. In current medical datasets, errors stemming from redundant reasoning paths account for a small proportion, hence the limited magnitude of accuracy improvement. Nevertheless, this still translates into a sustained 1-3 percentage point advantage in the AIT Index. Furthermore, as the weight of computational cost factors increases, SynapseRoute's advantage progressively widens, validating its resilience and adaptability under multiple constraints. In summary, SynapseRoute achieves optimal AIT performance across all evaluated weight settings, clearly demonstrating its capacity to further optimize inference costs while ensuring accuracy. This result proves that a dynamic routing strategy-based inference framework not only enhances model efficacy but also provides a more feasible and controllable solution for deploying large language models in real-world applications. The specific AIT results are presented in Table~\ref{tab:ait}.

\begin{table}[htbp]
    \centering
    \caption{AIT performance across different decision-making modes, with 95\% confidence intervals expressed as ± margins (estimated via bootstrapping).}
    \label{tab:ait}
    \begin{tabular}{m{5.5cm} >{\centering\arraybackslash}m{2.9cm} >{\centering\arraybackslash}m{2.9cm} >{\centering\arraybackslash}m{2.9cm}}
        \toprule
        \textbf{Scenario} & \textbf{Non-Thinking Mode} & \textbf{Thinking Mode} & \textbf{Dynamic Mode (SynapseRoute)} \\
        \midrule
        Accuracy First & 0.619 ± 0.025 & 0.834 ± 0.018 & 0.848 ± 0.017 \\
        Accuracy First + Cost Awareness & 0.662 ± 0.021 & 0.841 ± 0.016 & 0.859 ± 0.017 \\
        Inference Speed First & 0.745 ± 0.017 & 0.849 ± 0.012 & 0.874 ± 0.013 \\
        Balanced Strategy & 0.788 ± 0.013 & 0.859 ± 0.011 & 0.886 ± 0.011 \\
        Token Size Priority & 0.788 ± 0.014 & 0.863 ± 0.011 & 0.889 ± 0.010 \\
        \bottomrule
    \end{tabular}
\end{table}

\section*{Methods}

The methodology is structured into three key components: data preparation, the development pipeline of our routing algorithms, and the evaluation framework. These elements collectively underpin the design and validation of our proposed system. Figure~\ref{fig:pipeline}(c) provides a comprehensive overview of this entire process, illustrating the data flow through each stage, from preprocessing and model training to routing decision-making and performance evaluation, thereby offering a holistic view of the methodology.

\subsection*{Data}
\subsubsection*{Data Source}
We integrated four authoritative medical datasets: USMLE (the United States Medical Licensing Examination)~\citep{jin2021disease}, MedMCQA (focused on medical multiple-choice scenarios)~\citep{pal2022medmcqa}, PubMedQA (covering PubMed literature questions)~\citep{jin2019pubmedqa}, and CareQA (emphasizing fundamental medical knowledge assessment)~\citep{arias2025automatic}. These datasets were selected due to their high academic recognition, extensive supporting publications, and widespread download rates. We invited a professional medical team to select questions from the dataset that are more aligned with medical practice. To facilitate accurate evaluation, we excluded open-ended questions. Given computational resource limitations, we performed stratified sampling on the datasets, ensuring the sampled dataset structure maintained consistency with the original distribution. This resulted in a final research sample of 6,365 multiple-choice questions. We standardized all question options to uppercase letters A, B, C, D, E. For true/false/maybe question types in PubMedQA, we mapped "yes," "no," and "maybe" answers to A, B, and C, respectively, ensuring uniform formatting across multi-source data. We will open-source this labelled dataset.

\begin{figure}
    \centering
    \begin{subfigure}[b]{0.48\linewidth}
        \centering
        \includegraphics[width=\linewidth,trim={0 2cm 0 2cm},clip]{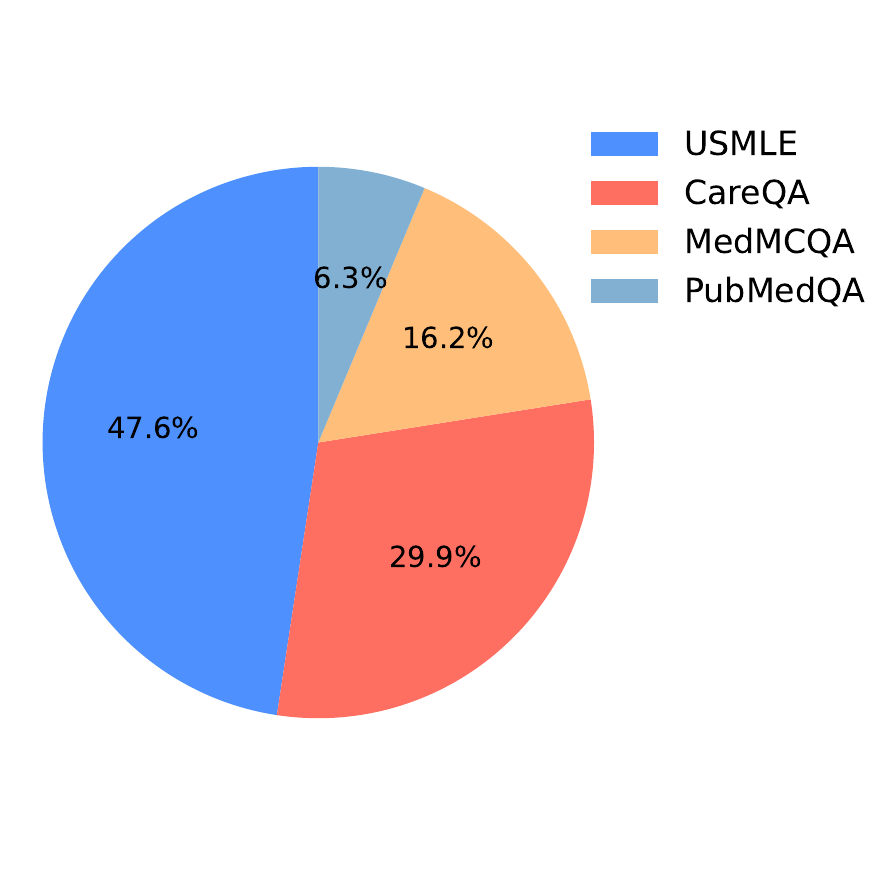}
        \caption{Data Source Distribution}
        \label{fig:data_source_distribution}
    \end{subfigure}
    \hfill
    \begin{subfigure}[b]{0.48\linewidth}
        \centering
        \includegraphics[width=\linewidth,trim={0 2cm 0 2cm},clip]{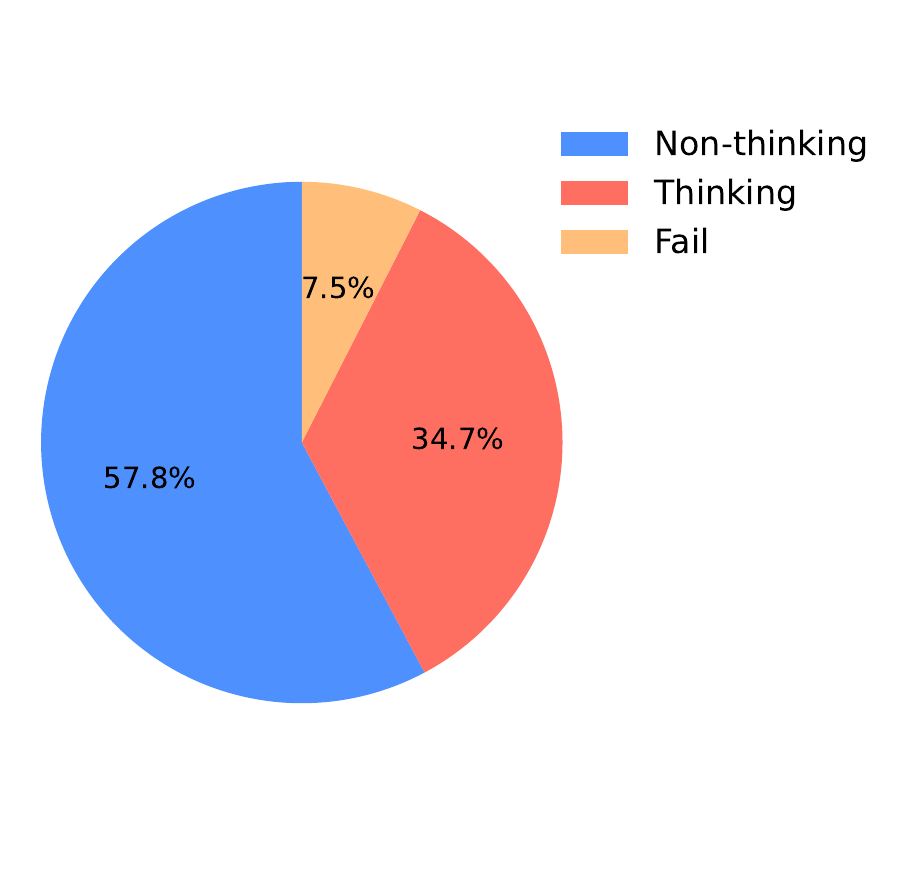}
        \caption{Ground Truth Label Distribution}
        \label{fig:true_label_distribution}
    \end{subfigure}
    \hfill
    \begin{subfigure}[b]{\linewidth}
        \centering
        \includegraphics[width=\linewidth]{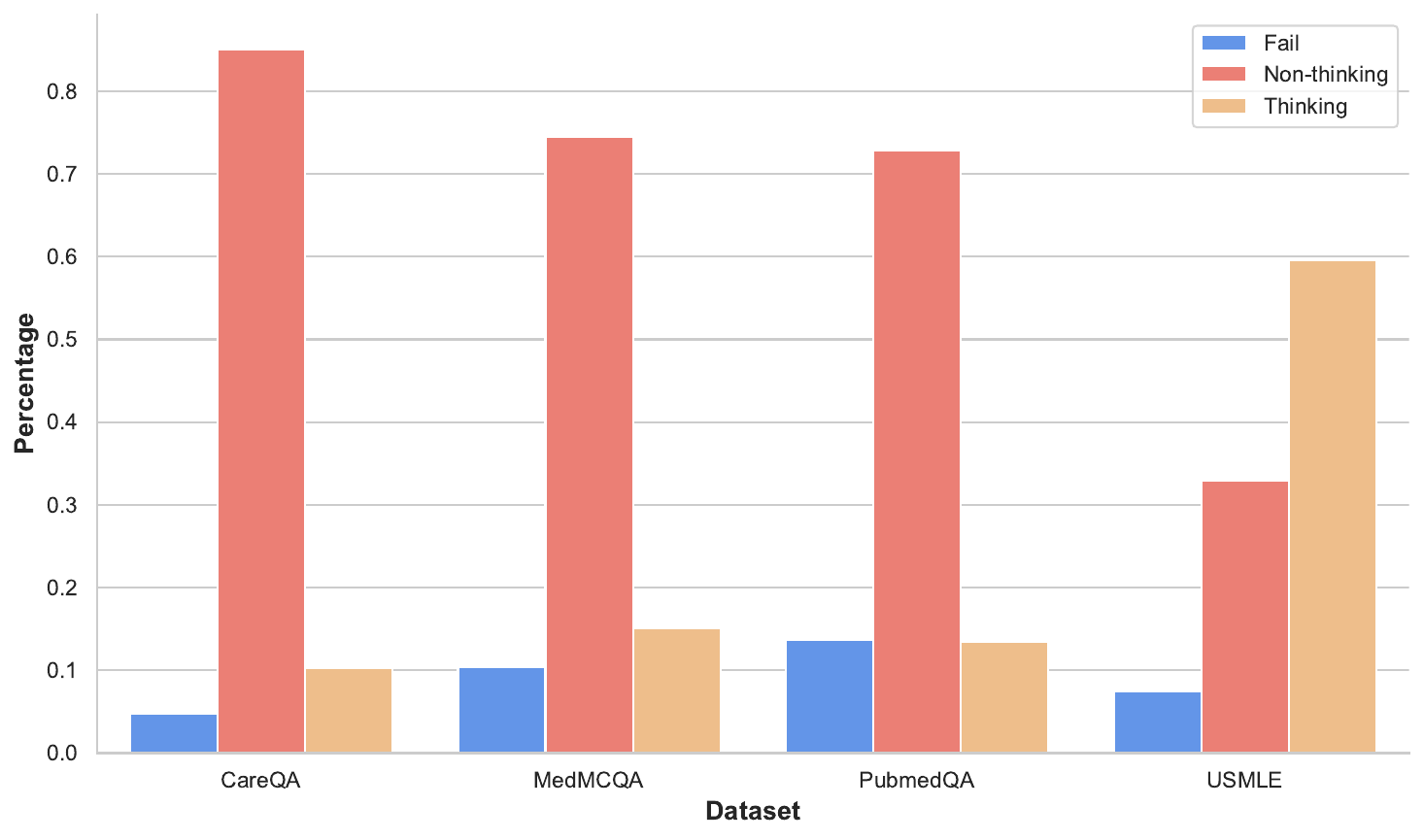}
        \caption{True Label Distribution By Data Source}
        \label{fig:true_label_distribution_by_data_source}
    \end{subfigure}
    \caption{Overview of Dataset Composition.}
    \label{fig:data_overview}
\end{figure}

As shown in the Figure~\ref{fig:data_source_distribution}, from a data composition perspective, USMLE data accounted for the highest proportion (47.6\%), with a higher percentage of comprehensive clinical reasoning questions. CareQA comprised 29.9\%, focusing on direct assessment of basic medical concepts. MedMCQA represented 16.2\%, concentrating on applied multiple-choice questions across various medical disciplines. PubMedQA made up 6.3\%, primarily consisting of factual questions related to literature. Together, these four datasets provide multi-level question coverage, ranging from fundamental knowledge points to complex clinical reasoning. 

\paragraph{USMLE}
The United States Medical Licensing Examination (USMLE)~\citep{jin2021disease} consists of questions drawn from professional medical board exams in the United States. It is widely regarded as a benchmark for evaluating medical knowledge and clinical reasoning.

\paragraph{PubMedQA}
PubMedQA~\citep{jin2019pubmedqa} is a question-answering dataset constructed from PubMed abstracts. The task involves answering research questions with one of three labels: yes, no, or maybe. For example: ``Do preoperative statins reduce atrial fibrillation after coronary artery bypass grafting?” The dataset includes 1,000 expert-annotated examples and is notable for being one of the first biomedical QA benchmarks requiring reasoning over scientific literature.

\paragraph{MedMCQA}
MedMCQA~\citep{pal2022medmcqa} is a large-scale multiple-choice question answering (MCQA) dataset designed to reflect real-world medical entrance exams, such as AIIMS and NEET PG. It comprises over 194,000 high-quality questions spanning 2,400 healthcare topics and 21 medical subjects. Each item includes a question, a correct answer, and distractor options. The dataset emphasizes complex reasoning and tests over 10 distinct cognitive abilities across diverse medical domains, with an average token length of 12.77.

\paragraph{CareQA}
CareQA~\citep{arias2025automatic} is derived from Spain's official Specialised Healthcare Training (MIR) exams, administered by the Spanish Ministry of Health. The multiple-choice version of the dataset includes 5,621 QA pairs, covering six major disciplines: medicine, nursing, biology, chemistry, psychology, and pharmacology. The questions are sourced from the 2020 to 2024 exam editions and represent a diverse, multilingual benchmark for medical QA.

\subsubsection*{Data Labelling}
For task complexity classification, we considered both manual and automated annotation strategies. Manual annotation, however, is not only labor-intensive and time-consuming but also suffers from high subjectivity, making it difficult to standardize what constitutes a task that "requires thinking." Moreover, as the capabilities of large language models (LLMs) continue to advance, some questions previously considered challenging can now be solved without explicitly invoking a reasoning or "thinking" mode.
To address these challenges, we adopt an automated annotation strategy, which offers greater scalability, consistency, and alignment with the practical goals of dual-mode inference. In this setting, model performance must account not only for accuracy but also for resource-related costs, such as token consumption and inference latency. Accordingly, we define a task as "thinking" or "non-thinking" based on three key metrics: answer accuracy, inference time, and token usage. This allows us to categorize question complexity in a way that reflects both cognitive demands and computational efficiency.
The annotation procedure is defined as follows:
\begin{itemize}
    \item Only one mode yields the correct answer:
    \begin{itemize}
        \item If only the \textit{thinking} mode produces the correct answer, the question is labeled as a \textbf{thinking question}.
        \item If only the \textit{non-thinking} mode produces the correct answer, the question is labeled as a \textbf{non-thinking question}.
    \end{itemize}
    \item Both modes yield the correct answer:
    \begin{itemize}
        \item If the \textit{thinking} mode has lower token usage and shorter inference latency, the question is labeled as a \textbf{thinking question}.
        \item Otherwise, if the \textit{non-thinking} mode is more efficient in terms of tokens and latency, the question is labeled as a \textbf{non-thinking question}.
    \end{itemize}
    \item Neither mode yields the correct answer:
    \begin{itemize}
        \item These questions are labeled as \textbf{fail} and excluded from the training set of the task complexity classifier.
    \end{itemize}
\end{itemize}

To implement this annotation strategy, we deployed the Qwen3-30B-a3b model for dual-mode inference on each medical question. Our method involved concatenating the question and options into a unified format ("system prompt + "question" + options") and calling both the thinking and non-thinking modes of the model to generate answers. We recorded the generated answer, token consumption (note: thinking mode token count includes its thought process), and inference time for each run, which served as the basis for annotation. Through this automated annotation process, we successfully classified the complexity types for 6,365 data entries, laying a solid data foundation for building a supervised learning dynamic routing model. This strategy not only avoids biases from subjective judgment but also significantly improves annotation efficiency and consistency, demonstrating strong generalization and scalability for broad application in other dual-mode task scenarios.

\subsubsection*{Exploratory Data Analysis}

After completing the automated annotation of all medical data, we conducted a statistical analysis of the distribution of question types. The results show that non-thinking questions constituted 57.8\%, thinking questions comprised 34.7\%, and 7.5\% of questions were labeled as "fail", meaning the model failed to provide a correct answer regardless of whether thinking mode was enabled.

Analyzing by data source, different datasets exhibited clear distinctions in question difficulty and design focus. In Figure~\ref{fig:true_label_distribution}, we present the distribution of the annotated data results, as well as the distribution of results by source. The USMLE dataset had the highest proportion of thinking questions, nearly 60\%, indicating that questions in this dataset typically involve strong logical reasoning and professional judgment. In contrast, CareQA showed less than 10\% thinking questions, with over 80\% non-thinking questions, reflecting its emphasis on direct retrieval of basic knowledge and relatively lower inference complexity. Both MedMCQA and PubMedQA datasets had around 72\% non-thinking questions. Notably, PubMedQA had a higher proportion of "fail" questions, reaching 13\%. This phenomenon might be related to PubMedQA's task format, which consists of yes/no/maybe tri-classification questions that inherently carry more semantic uncertainty, making them more prone to model errors. As exemplified in Table~\ref{tab:examples_of_thinking_questions}, non-thinking questions are typically more direct, close to knowledge recall and factual judgments. In contrast, thinking questions are generally longer, contain more information, and necessitate multi-step reasoning.

\begin{table}[htbp]
    \centering
    \caption{Representative examples of thinking questions and non-thinking questions, sampled from benchmark datasets.}
    \begin{tabular}{m{0.83\linewidth}m{0.12\linewidth}}
        \toprule
        \textbf{Question} & \textbf{Label}  \\
        \midrule
What type of stem cells do hematopoietic progenitors belong to? & non-thinking \\
A 45-year-old woman comes to see you for a second opinion regarding an upcoming surgery for pancreatic insulinoma. While taking a surgical history, she tells you she previously had a pituitary tumor resected. For which additional neoplasms might you consider testing her? &  thinking \\
A neuralgia with trigger zones in the oropharynx and pain in the ear pharynx, nasopharynx, tonsils and posterior tongue is most likely: &  non-thinking \\
For a certain chemical reaction, it is found that, in the temperature range between 100$^{\circ}\mathrm{C}$ and 300$^{\circ}\mathrm{C}$, both the standard Gibbs reaction energy, $\Delta\mathrm{rG}^\circ$, and the standard reaction enthalpy, $\Delta\mathrm{rH}^\circ$, are positive quantities. Which of the following pairs of equilibrium constants is compatible with the previous data?: &  thinking \\
        \bottomrule
    \end{tabular}
    \label{tab:examples_of_thinking_questions}
\end{table}

We also compared the inference time and token costs between using exclusively thinking mode and exclusively non-thinking mode for all questions in Table~\ref{tab:resource_consumption}. Thinking mode significantly outperformed non-thinking mode in terms of inference resource consumption. Specifically, thinking mode had an average inference token count of 799.31 tokens and an average inference duration of 17.24 seconds. In contrast, non-thinking mode averaged only 5 inference tokens and 1.25 seconds of inference duration. This significant difference was also prominent in quantile analysis: the 50th percentile for thinking mode output was 618 inference tokens and 13.19 seconds of inference duration, far exceeding non-thinking mode's consistent 5 tokens and 0.85 seconds at the same percentile. It is worth noting that since this study used only multiple-choice questions, and non-thinking mode doesn't require multi-round inference or a thinking process from the model, its output is often a fixed-length answer token (e.g., "B" or "C"), keeping token count largely constant. Consequently, inference duration also remains low. The thinking mode's inference process, however, includes the generation of the model's thought process, leading to a substantial increase in token count and time cost. Taking extreme values as an example, some thinking mode inferences reached as high as 8,277 tokens and exceeded 3 minutes (184.41 seconds) in duration, whereas the maximum inference duration for non-thinking mode was only 80.95 seconds. These statistics clearly demonstrate the vast difference in inference resource consumption between thinking mode and non-thinking mode within the same foundational model. We further illustrate the significant potential for cost optimization by accurately identifying question type and bypassing unnecessary thinking processes while maintaining accuracy.

\begin{table}[htbp]
    \centering
    \caption{Comparison of Resource Consumption by Different Inference Mode}
    \begin{tabular}{m{3.5cm}|m{2cm}|m{3cm}|m{2cm}|m{3cm}}
        \toprule
        &  \multicolumn{2}{c|}{\textbf{Thinking Mode}}  & \multicolumn{2}{c}{\textbf{Non-Thinking Mode}}  \\
        \hline
        & Token Size & Inference Time (s) & Token Size & Inference Time (s) \\
        \midrule
       \textbf{Count} & \multicolumn{4}{c}{6365}	\\
       \hline
\textbf{Mean} & 799.31 & 17.24 & 5.00 & 1.25 \\
\textbf{Standard Deviation} &	581.11&	13.60&	0.00	&1.90 \\
\textbf{Min} & 128.00	&2.70	&5.00	&0.62 \\
\textbf{25\%} & 410.00	&8.60&	5.00	&0.79 \\
\textbf{50\%} & 618.00	&13.19	&5.00	&0.85 \\
\textbf{75\%} & 1017.00	&21.00&	5.00&	0.95 \\
\textbf{Max} & 8277.00&	184.41&	5.00&	80.95 \\
        \bottomrule
    \end{tabular}
    \label{tab:resource_consumption}
\end{table}

\subsection*{Routing Algorithm}

Following pre-defined annotation rules, we used an automated script to precisely categorize the raw dataset into three classes: "thinking," "non-thinking," and "fail." The "fail" label was assigned to instances where the LLM failed to answer correctly in both operational modes. To enhance the focus and efficiency of training classification algorithm, we excluded "fail" category data from the training set due to its weaker relevance to our core optimization objective. Statistical analysis revealed a significant data distribution imbalance: "non-thinking" questions far outnumbered "thinking" questions. Specifically, "non-thinking" questions constituted 57\% of the total dataset, "thinking" questions made up 34.75\%, and "fail" questions accounted for 7.51\%. To mitigate the risk of model prediction bias caused by this sample imbalance, we employed stratified sampling for a scientific train-test split, effectively preserving the original sample distribution characteristics. After this division, the training set contained 4,710 data points, and the test set contained 1,177 data points, with specific proportions detailed in Table~\ref{tab:data_distribution} below.

\begin{table}[htbp]
    \centering
    \caption{Data Distribution in Training and Testing}
    \begin{tabular}{m{1.3cm}|m{2cm}|m{2.5cm}|m{2.5cm}|m{1.8cm}|m{2.5cm}}
        \toprule
\textbf{Datasets}	& \textbf{Thinking Question (\%)} &	\textbf{Non-thinking Question (\%)} &	\textbf{Fail Question (\%)} & 	\textbf{Total Count} &\textbf{Total Count (Binary)} \\
\midrule
\textbf{Total}	&34.74&	57.75&	7.51	&6365	&5887\\
\textbf{Train} &	34.74&	57.76&	7.5	&5092&	4710\\
\textbf{Test}	&34.72&	57.74&	7.54	&1273	&1177\\
        \bottomrule
    \end{tabular}
    \label{tab:data_distribution}
\end{table}

For routing algorithm selection, we considered performance, training cost, and deployment efficiency, exploring two paths to address the binary classification problem: traditional machine learning methods and Large Language Model fine-tuning methods. Each approach offers distinct advantages: traditional machine learning is simple to implement, computationally inexpensive, and highly interpretable, especially suitable for small to medium-scale datasets. In contrast, LLM fine-tuning boasts powerful representation capabilities and cross-task transferability, excelling in more complex and semantically rich tasks. Our goal was to compare the practical performance of these two technical routes on our task to provide data-driven support for the final model selection.

\begin{figure}
    \centering
    \begin{subfigure}[b]{0.48\linewidth}
        \centering
        \includegraphics[width=\linewidth]{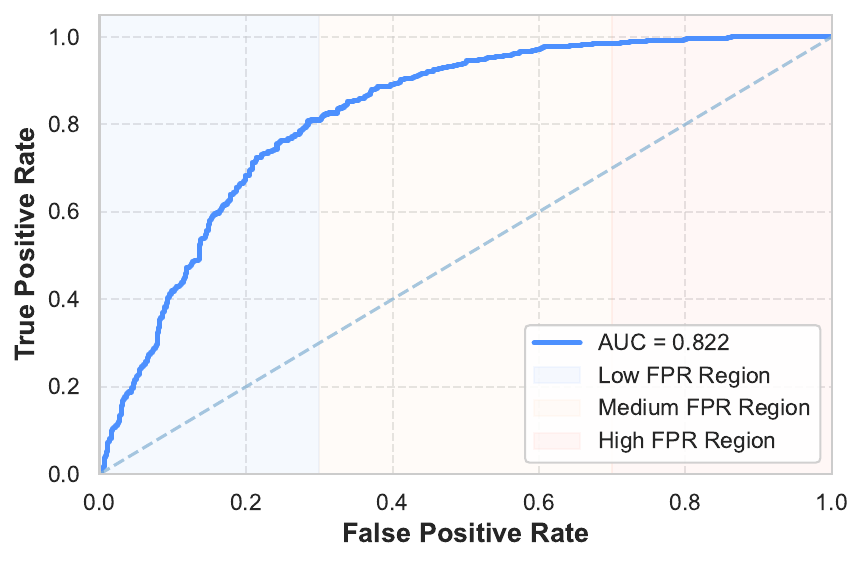}
        \caption{ROC Curve for Logistic Regression in Test Set}
    \label{fig:auc_logistic_regression}
    \end{subfigure}
    \hfill
    \begin{subfigure}[b]{0.51\linewidth}
        \centering
        \includegraphics[width=\linewidth]{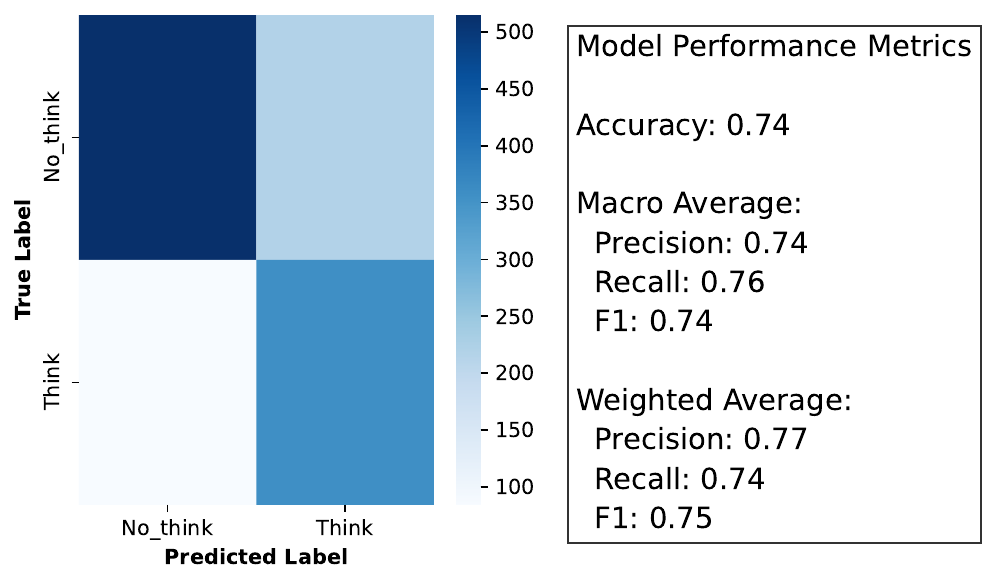}
        \caption{Confusion Matrix of Logistic Regression Model}
        \label{fig:logistic_regression_confusion_f1}
    \end{subfigure}
    \hfill
    \begin{subfigure}[b]{\linewidth}
        \centering
        \includegraphics[width=\linewidth]{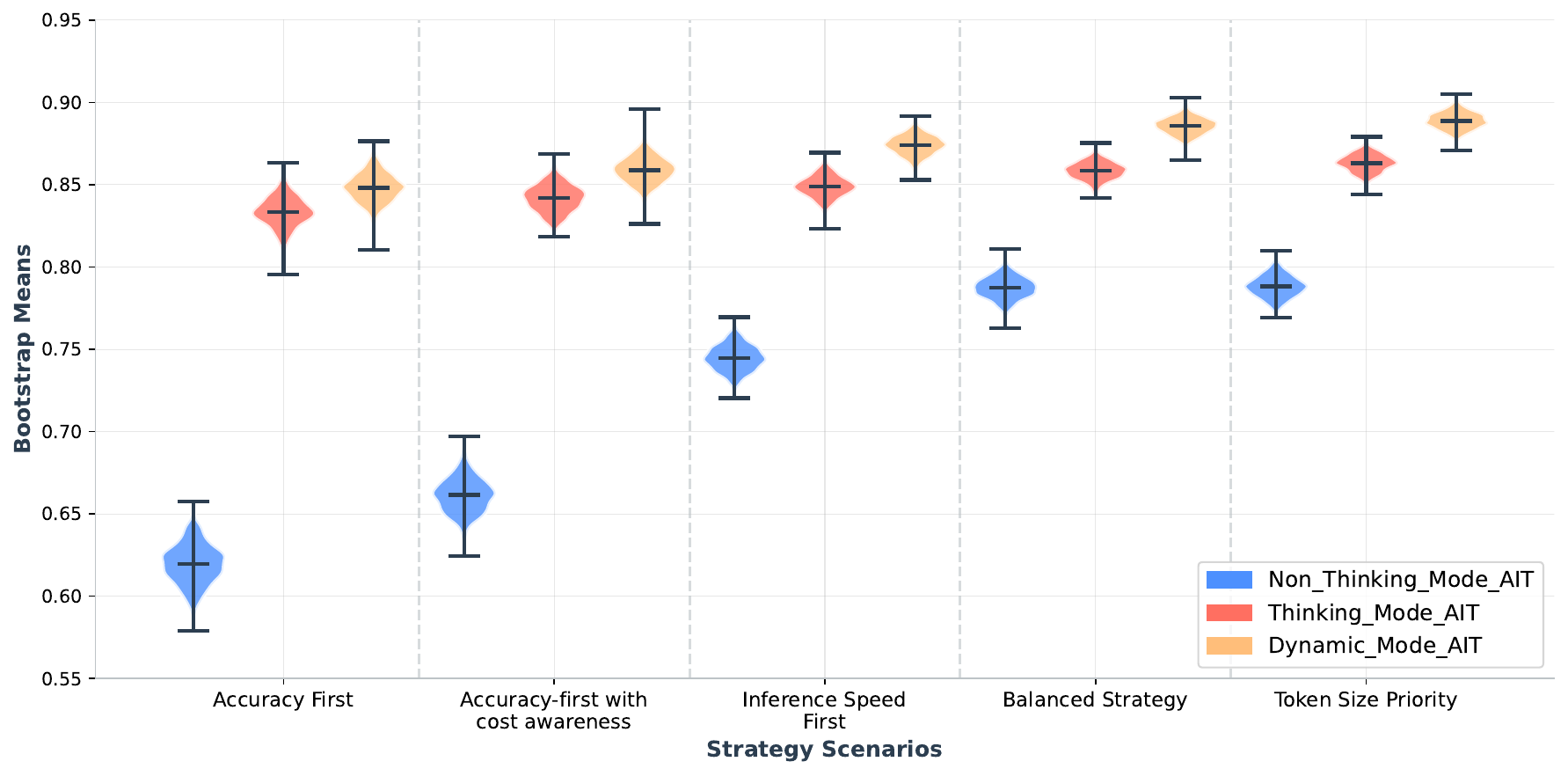}
        \caption{Comparison of Bootstrap Sample Means across Different Scenarios}
        \label{fig:ait_by_scenario_0605}
    \end{subfigure}
    \caption{Quantitative Results of Synapseroute Framework.
    }
    \label{fig:quat_results}
\end{figure}

For the traditional machine learning approach, we began by vectorizing the raw text data using an open-source text embedding model. Specifically, we used the BAAI/bge-large-en-v1.5 model from the Hugging Face platform to extract embeddings for all questions, converting text into fixed-dimensional vector representations. The primary benefit is that pre-trained embedding models effectively capture semantic information, enabling subsequent classification models to learn and generalize better at a semantic level. Vectorization also significantly improves comparability and transferability between different models, facilitating unified modeling and evaluation. After obtaining the vector representations, we sequentially experimented with several mainstream binary classification models, including Logistic Regression, XGBoost, LightGBM, and Random Forest. For each model type, we performed an extensive and meticulous hyperparameter tuning process, leveraging a comprehensive grid search with cross-validation. This allowed us to rigorously explore a broad and finely-grained hyperparameter space to identify the best-performing model configuration for the task. On the test set, the AUC performance of the various models was as follows: Logistic Regression achieved an AUC of 0.82, LightGBM reached 0.81, XGBoost also scored 0.82, while both Random Forest and CatBoost obtained 0.80. Figure~\ref{fig:auc_logistic_regression} displays the ROC curve for Logistic Regression on the test set. Considering both model performance and complexity, we ultimately selected Logistic Regression as the optimal model within the traditional machine learning path. In the inference phase, as Logistic Regression is a probabilistic output model, we further aimed to maximize the F1 score, selecting an optimal threshold for final class prediction.

We also conducted experiments using parameter-efficient fine-tuning (PEFT) on Qwen3-4B, a 4-billion-parameter model from the Qwen series known for its strong general language understanding capabilities. Directly fine-tuning such a large model entirely is computationally expensive and prone to catastrophic forgetting. Therefore, we employed LoRA (Low-Rank Adaptation)~\citep{hu2022lora} for fine-tuning. LoRA is a lightweight PEFT technique that works by introducing additional low-rank adaptation modules while keeping the original pre-trained model parameters frozen. 
For the base model selection, considering computational resources and cost efficiency, we performed LoRA fine-tuning on the Qwen3-4B model. After conducting multiple rounds of thorough hyperparameter tuning and extensive experimentation, we selected the configuration that consistently yielded the best performance. Specifically, we set the LoRA parameters to lora\_rank = 32 and lora\_alpha = 64, with a learning rate of 0.00002. This carefully chosen setup enabled the fine-tuned model to achieve a test accuracy of 0.73, demonstrating the effectiveness of our tuning strategy and the suitability of these parameter choices for the task.

After training all models, we systematically evaluated them on the test set, using accuracy and Area Under the Receiver Operating Characteristic Curve (AUC), a representative metric for class-imbalanced problems, as primary evaluation indicators. The Logistic Regression model, based on traditional machine learning methods, performed best, achieving an AUC of 0.82 and an accuracy of 0.74. This not only outperformed other embedded classification models like XGBoost, LightGBM, and Random Forest but also slightly surpassed the Qwen3-4B LLM fine-tuned with LoRA (which had an accuracy of approximately 0.73). 
As shown in Figure~\ref{fig:logistic_regression_confusion_f1}, the confusion matrix analysis indicates that the Logistic Regression model achieved a macro-average F1 score of 0.74 and a weighted-average F1 score of 0.75, both of which are approximately 1\% higher than those of the LoRA-tuned Qwen3-4B model.
This indicates its more balanced and consistent advantage in overall classification performance. Experimental analysis indicates that for the current medium-scale medical question-answering corpus, high-quality features extracted using semantic embeddings enable linear models like Logistic Regression to effectively distinguish between "thinking" and "non-thinking" questions. Compared to large language models, which have vast parameter counts and high training costs, Logistic Regression offers advantages such as a simple structure, fast inference, lightweight deployment, and strong interpretability. These characteristics align more closely with our strict requirements for model response speed and resource utilization within the SynapseRoute framework. Therefore, we ultimately selected the Logistic Regression model as the core decision component in SynapseRoute for routing medical questions, that is, determining whether a question needs to enable "thinking" mode. This design not only ensures a strong balance between system performance and cost but also lays an efficient and controllable foundation for subsequent modular deployment and LLM inference routing.
\begin{table}[htbp]
    \centering
    \caption{Metric Definition Table}
    \begin{tabular}{m{0.17\linewidth}|m{0.15\linewidth}|m{0.58\linewidth}}
        \toprule
\textbf{Metric Name} & \textbf{Value Range} & \textbf{Meaning of Values} \\
\midrule
Accuracy & 0/1 & 0 indicates incorrect answer, 1 indicates correct answer \\
Inference Time & [0,1] & The closer to 1, the less inference time required \\
Token & [0,1] & The closer to 1, the fewer tokens output \\
        \bottomrule
    \end{tabular}
    \label{tab:metric_def}
\end{table}
\subsection*{Evaluation Framework}

\subsubsection*{Single-Dimensional Performance Metrics}

In the medical field, accuracy is the primary core metric of concern; the reliability of diagnostic results is paramount, as errors at any stage can severely impact patient health. Consequently, the accuracy of model responses holds a dominant position in this study's evaluation system. Beyond accuracy, this research employs Macro-averaged and Weighted-averaged forms of precision, recall, and F1-score to multi-dimensionally assess the model's overall performance across different answer options. Specifically, Macro-averaged metrics provide a fair reflection of the model's performance across all categories by equally considering the performance of different sample classes. In contrast, weighted-averaged metrics are calculated based on the true sample counts of each class, more accurately mapping the model's performance efficacy within the actual data distribution. The combined evaluation results from these metrics offer a comprehensive and objective insight into the model's classification capability and generalization level in medical multiple-choice scenarios.

\subsubsection*{Cross-Dimensional Joint Evaluation Metrics: AIT Index}

In this study, to comprehensively evaluate models by integrating both accuracy and cost factors, we developed the AIT Index (Accuracy, Inference Time, Token Count). This index serves as a holistic metric for assessing the trade-off between model accuracy and cost. It not only focuses on the correctness of the model's output but also incorporates the user experience dimension (model response time) and the practical usage cost dimension (token consumption, which is directly correlated with cost). Through quantitative analysis, the AIT Index enables a comprehensive and multi-faceted evaluation of the model's real-world application experience. Compared to single performance metrics, the AIT Index more realistically reflects the balance between efficiency and economic viability in actual business scenarios, providing a scientific basis for model optimization and deployment decisions.
\begin{table}[htbp]
    \centering
    \caption{Scenario - Weighting Strategy Specification Table}
    \begin{tabular}{m{3cm}|m{4cm}|m{7.7cm}}
        \toprule
\textbf{Scenario Name} & \textbf{Weight Values} & \textbf{Scenario Meaning} \\
\midrule
Accuracy First &  a=0.9, b=0.05, c=0.05 &  Prioritizes accuracy with a high weight of 0.9, while setting inference time and token cost weights to 0.05 each. \\
\hline
Accuracy-first with cost awareness & a=0.8, b=0.05, c=0.15 & Maintains a high accuracy weight of 0.8 while increasing the token cost weight to 0.15 for cost consideration. \\
\hline
Inference Speed First & a=0.6, b=0.3, c=0.1 & Emphasizes response efficiency, with an inference time weight of 0.3, accuracy weight of 0.6, and token cost weight of 0.1. \\
\hline
Balanced Strategy & a=0.5, b=0.25, c=0.25 & Adopts a balanced approach, assigning weights of 0.5 to accuracy, 0.25 to inference time cost, and 0.25 to token cost. \\
\hline
Token Size Priority & a=0.5, b=0.1, c=0.4 & Focuses on token cost with a weight of 0.4 while maintaining an accuracy weight of 0.5 and a low inference time weight of 0.1. \\
        \bottomrule
    \end{tabular}
    \label{tab:weighting_strategy_specification}
\end{table}
The AIT Index is calculated using the formula: $\mathrm{AIT}=a\times A+b\times I+c\times T$. Here, A represents the model accuracy, I denotes the standardized inference time, and T signifies the standardized token consumption (token size). The parameters a, b, and c are weight coefficients for these three components, which can be dynamically adjusted based on the core evaluation priorities of different application scenarios (e.g., performance-first or cost-sensitive). This allows the index calculation to more closely align with actual business needs for model evaluation. Through this tunable linear weighting mechanism, the AIT Index can flexibly adapt to diverse evaluation objectives, enabling personalized quantitative analysis of model performance and cost.

At the metric construction level, accuracy is a binary variable (0 or 1) indicating whether the model's output is correct. For inference time and token consumption, which have different units, this study employs Min-Max Normalization (Equation~\ref{eq:min-max_normalization}) to linearly scale their feature values to the [0,1] range. This ensures numerical compatibility with the accuracy metric for subsequent calculations. Importantly, since inference time and token consumption are negative cost indicators (larger values imply worse performance), we apply an inversion transformation (1-x) to these normalized metrics. This converts them into positive indicators, where a larger value signifies lower cost and higher efficiency. This processing preserves the relative distribution characteristics of the original data and eliminates dimensional differences that could interfere with comprehensive evaluation, allowing multi-dimensional metrics to be linearly weighted and aggregated on the same polarity and scale.  Table~\ref{tab:metric_def} illustrates the value ranges and meanings of each parameter.

\begin{equation}
    \label{eq:min-max_normalization}
    x' = \frac{x-\min(X)}{\max(X) - \min(X)},
\end{equation}

In this study, given the high dependence on accuracy in medical scenarios, where diagnostic reliability is paramount, and any error can severely impact patient health, we consistently positioned accuracy as the core dominant factor within the AIT Index. Its weight is always set to be greater than 0.5, ensuring that the model evaluation prioritizes the reliability of results. Simultaneously, to address the varied demands for model inference speed and computational resource consumption across different business scenarios, we designed five typical weighting strategies, establishing a flexible and adjustable AIT comparison system. Specifically, the "Accuracy First" scenario places extreme emphasis on the correctness of model output, assigning a high weight of 0.9 to accuracy, with minimal consideration for inference time and token cost. "Accuracy-first with Cost Awareness" maintains a high accuracy weight while moderately increasing sensitivity to token consumption, suitable for deployment scenarios that require balancing stability with expenditure. The "Inference Speed First" and "Token Size Priority" strategies respectively prioritize response timeliness and resource costs, making them more appropriate for online systems highly sensitive to interaction efficiency or API costs. Finally, the "Balanced Strategy" allocates more equitable weights to all three factors, suitable for general deployment needs, aiming for a reasonable compromise between performance and cost. By constructing this parameter-adjustable AIT evaluation framework with multiple assessment dimensions, we can achieve personalized comparative and quantitative analysis of model efficacy across diverse application scenarios. This effectively supports model selection, deployment strategy formulation, and long-term optimization pathway planning, thereby not only enhancing the decision-making value of the evaluation but also providing scientific support for building medical question-answering systems that meticulously balance accuracy, efficiency, and cost control. More detailed information is provided in Table~\ref{tab:weighting_strategy_specification}.

\section*{Discussion}

Since the advent of LLMs, inference cost has become a critical concern alongside response correctness, especially in resource-constrained environments where it poses a significant barrier to adoption~\citep{strubell2020energy,kaplan2020scaling,zheng2024dape}. Models of different sizes exhibit clear discrepancies in application effectiveness and cost. Researchers have explored various routing architectures to collectively reduce model usage costs~\cite{sun2021hiabp,sun2024hierarchical}. Chen et al.~\citep{chen2023frugalgpt} proposed a framework combining prompt adaptation and model cascading: a cheaper model first generates candidate answers, which an expensive model then refines, achieving high accuracy with substantial cost reduction. Kolawole et al.~\citep{kolawole2024revisiting} utilized a rule-based adaptive inference strategy, building a cascade from smaller to larger models, automatically escalating to a larger model when smaller model outputs are inconsistent. The fundamental idea behind these related studies is to leverage the cost-effectiveness of smaller models for simpler questions via cascading, while reserving larger models for problems that smaller models struggle with, thereby lowering overall inference costs.  However, these approaches do not explicitly define problem classification. Moreover, they often involve deploying multiple separate models, resulting in increased deployment complexity and higher operational costs.

With the release of DeepSeek-R1, reasoning models have gained increased attention due to their white-box thinking processes, complete inference chains, and higher accuracy in complex problems. However, the token and time costs of reasoning models are significantly higher than those of chat-optimized models. A new research challenge is identifying the scenarios in which to invoke the "thinking mode". That is, determining which problems are better addressed by reasoning models to reduce overall model costs. Regarding problem classification, Marinelli et al.~\citep{marinelli2025harnessing} proposed using "Number of Thoughts (NofT)" as a metric for problem complexity, routing problems to appropriate models based on a threshold. However, question-answer pairs in different domains might exhibit varying tendencies for this metric, and the NofT output by the model itself could also be biased. In the medical domain, data professionalism and specificity are even more pronounced. Zhang et al.~\citep{yang2023one} introduced the LLM-Synergy framework, employing weighted voting and a cluster-based dynamic model selection strategy for medical question answering tasks, demonstrating significant improvement when using multiple models compared to a single one. Kim et al.~\citep{kim2024mdagents} utilized a "moderator" LLM to classify the complexity of input queries before routing them to different intelligent agents for responses, enhancing model performance in medical tasks. However, the effectiveness of this classification task depends on the performance of the LLM itself, making it challenging to further verify the accuracy of its complex problem judgments. Furthermore, deploying multiple models and constructing agents significantly increases costs.

In general, existing methods primarily focus on routing decisions among multiple models, inherently relying on the simultaneous deployment of several LLMs to strike a balance between cost and performance. Although effective to some extent, these strategies inevitably introduce greater system complexity and deployment costs. The emergence of dual-mode LLMs offers a promising avenue for a simpler, lower-cost solution to this problem. For instance, the Qwen3 open-source model integrates "thinking mode" and "non-thinking mode" within a single model, significantly reducing the resource overhead required for multi-model deployment. This model supports two implementation methods to switch thinking mode: one by modifying parameters (e.g., setting \texttt{enable\_thinking=True/False}) to enable or disable thinking mode; the other by prompt control, adding \texttt{/think} or \texttt{/no\_think} to the system prompt to switch modes. This mode fusion capability provides the technical foundation for more efficient task routing. However, current dual-mode models still rely on explicit user control or manual rule triggers, lacking a universal, automated judgment mechanism for intelligent switching based on problem complexity. Therefore, there is an urgent need for an automated routing framework that can achieve intelligent mode selection within a unified model architecture to truly unleash the potential of dual-mode models. In this context, this study proposes the SynapseRoute framework. We aim to precisely judge the complexity of medical queries through an automated labeling strategy for medical question-answering data, thereby building a supervised learning model to automatically activate or deactivate the model's thinking mode. This method significantly reduces inference latency and token cost while ensuring answer accuracy, offering a new solution for the efficient deployment of LLMs in the medical field and laying the groundwork for further exploration of dual-mode LLM applications.

This study further validated SynapseRoute in the medical domain. To this end, we developed a systematic evaluation framework and comparatively analyzed SynapseRoute's performance against two single modes (thinking and non-thinking). Experimental results demonstrate that SynapseRoute not only offers significant improvement in accuracy but also achieves a superior balance between performance and cost. We believe that integrating multiple inference modes within a single LLM, combined with intelligent routing strategies, is a promising future development direction for multi-mode LLM architectures. In practical applications, the SynapseRoute framework can quickly identify question types, effectively reducing inference costs and improving user interaction experience while ensuring answer accuracy. 

Despite the positive outcomes of this study, several limitations remain. Firstly, while the four public medical datasets (MedQA, MedMCQA, PubMedQA, CareQA) are representative, their coverage of clinical sub-domains is not yet comprehensive. This might limit their ability to fully reflect SynapseRoute's applicability across all types of medical tasks.
Furthermore, this research has only conducted experiments in the medical domain, and the generalization capability of this framework mechanism in other highly specialized fields, such as law or finance, remains unverified. Given the strong professionalism and structured nature of medical data, the transferability of this strategy to other domains requires further research and empirical support.

\bibliography{reference}
\bibliographystyle{apalike}

\end{document}